\UseRawInputEncoding
\documentclass[journal]{IEEEtran}
\usepackage[colorlinks,urlcolor=blue,linkcolor=blue,citecolor=blue]{hyperref}
\usepackage{multirow}
\usepackage{graphicx}
\usepackage{mathtools}
\usepackage{commath}
\usepackage{amsmath}
\usepackage{graphicx}
\usepackage[linesnumbered,ruled,vlined]{algorithm2e}
\usepackage{epstopdf}
\usepackage{multirow}
\usepackage{array}
\usepackage{pgfplots}
\usepackage{graphicx}
\usepackage{amssymb} 
\begin{document}
\title{Towards Robust and Accurate Myoelectric Controller Design based on Multi-objective Optimization using Evolutionary Computation}
\author{Ahmed Aqeel Shaikh, Anand Kumar Mukhopadhyay, Soumyajit Poddar, and Suman Samui 
\thanks{A. A. Shaikh is with School of Electrical and Computer Engineering, Georgia Institute of Technology Atlanta, USA (e-mail: ashaikh65@gatech.edu), A.K. Mukhopadhyay is with MathWorks India Private Limited (e-mail: anand.mukhopadhyay@gmail.com), S. Poddar is with Department of Electronics and Communication Engineering, Indian Institute Information Technology Guwahati (poddar18@gmail.com), and S. Samui is with Department of Electronics and Communication Engineering, National Institute of Technology Durgapur, West Bengal -  713209, India (e-mail: samuisuman@gmail.com).}}

%%\markboth{Journal of \LaTeX\ Class Files, Vol. 14, No. 8, August 2015}
%\markboth{}
\maketitle
\begin{abstract}
Myoelectric pattern recognition is one of the important aspects in the design of the control strategy for various applications including upper-limb prostheses and bio-robotic hand movement systems. The current work has proposed an approach to design an energy-efficient EMG-based controller by considering a kernelized SVM classifier for decoding the information of surface electromyography (sEMG) signals to infer the underlying muscle movements. In order to achieve the optimized performance of the EMG-based controller, our main strategy of classifier design is to reduce the false movements of the overall system (when the EMG-based controller is at the `Rest' position). To this end, we have formulated the training algorithm of the proposed supervised learning system as a general constrained multi-objective optimization problem. An elitist multi-objective evolutionary algorithm $-$ the non-dominated sorting genetic algorithm II (NSGA-II) has been used to tune the hyperparameters of SVM. We have presented the experimental results by performing the experiments on a dataset consisting of the sEMG signals collected from eleven subjects at five different upper limb positions. Furthermore, the performance of the trained models based on the two-objective metrics, namely classification accuracy, and false-negative have been evaluated on two different test sets to examine the generalization capability of the proposed training approach while implementing limb-position invariant EMG classification. It is evident from the presented result that the proposed approach provides much more flexibility to the designer in selecting the parameters of the classifier to optimize the energy efficiency of the EMG-based controller.
\end{abstract}
\IEEEpeerreviewmaketitle
\section{Introduction}
\IEEEPARstart{T}{here} has been a considerable amount of progress in the research and development of artificial hands controlled by electromyography signals \cite{gopura2016developments}. The artificial hand is a kind of Human-Computer Interaction (HCI) system which performs hand movements pertaining to wrist, grip, and fingers \cite{khushaba2014towards}. The design of this type of assistive technology requires a proper blending of signal processing with computational intelligence. Moreover, it has to satisfy various conflicting optimization goals at multiple stages of its operation towards achieving the best possible performance in all aspects. With the advancement in machine learning (ML) algorithms, smart controllers are used for developing robust and efficient HCI systems \cite{oskoei2007myoelectric}. The controller is analogous to the brain of the entire system, which is supposed to map the EMG signals to the user's intended hand movement accurately. The focus of the current work is to develop a supervised learning strategy to optimize the performance of the controller for a robust HCI system for hand movement classification \cite{subasi2013classification}\cite{rabin2020classification}\cite{reynoso2014controller}.

A complete system of an artificial hand movement classification using surface electromyography (sEMG) signal consists of several units composed of electronic and mechanical sub-components as shown in Fig. 1. Firstly, the raw sEMG signal is pre-processed and filtered to remove the unwanted noise frequencies and enhance the signal strength. Next, useful distinguishable information is extracted from the filtered signal, more commonly known as the feature extraction process. Following which, the predictor variables (feature vectors) are fed to a classifier which is an EMG-based controller (EBC). The EBC is used to control an exoskeleton or a prosthetic arm for a healthy or specially-abled person respectively, although the latter case is more cumbersome compared to the former. There are multiple ways of developing an EBC for upper-limb prostheses \cite{gopura2013recent}. In some cases of an advanced prosthesis, an additional controller is also used for incorporating more functionalities in the prosthetic/exoskeleton device \cite{parajuli2019real}. One such example is the cyborg drummer where the additional controller is activated based on the background music to facilitate an extra functionality for a superior drumming experience \cite{weinberg2020wear}. The design of the controller is determined by the type of artificial intelligence algorithm used to generate the controller output and the specification of the artificial mechanical hand system required for the application. 
%------------------------------------------------------------
\begin{figure}[h]
  \centering
  \includegraphics[scale = 0.38]{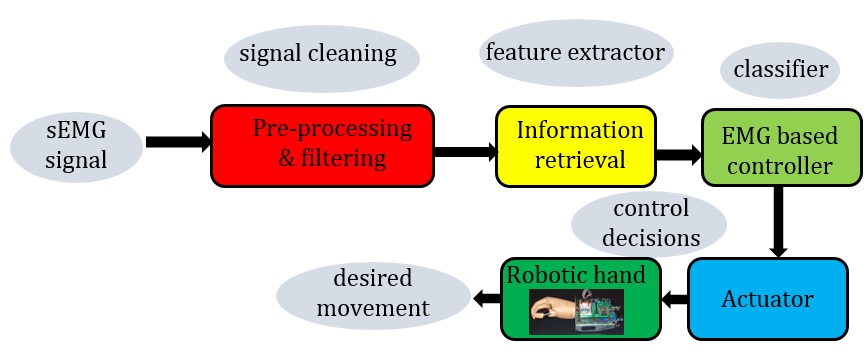}
  \caption{Robotic hand movement system based on sEMG signal classification.}
  \label{fig1}
\end{figure}
%------------------------------------------------------------
The output of the EBC is a control decision signal for the desired movement of the subsequent bio-robotic hand steered by a suitable actuator \cite{gopura2013recent}. The actuator predicts the angle of rotation to accomplish a desired classification hand movement. An improved dual-mode actuator, consisting of fluidic and tendon-driven mechanisms, was used in \cite{li2021dual}, which helps in the effective increase of the bending angle with multiple degrees of freedom \cite{su2021deep}. 

The main challenges in moving towards a cost-effective and power-efficient wearable prosthetic hand control system are the time and expertise required to optimize at all the stages in the workflow starting from data collection to deployment \cite{brunelli2015low}. In \cite{guzman2017very}, the authors have developed an event-based method for low-power sEMG acquisition systems, i.e., during the recording of the sEMG signal. Approximate computing for machine learning models \cite{masadah2021}\cite{castro2020} and bio-inspired classifiers, such as Spiking Neural Networks (SNN), were found to be effective in reducing the computational complexity of the classifier, but with a loss in accuracy \cite{mukhopadhyay2018classification}.

It is worth mentioning that it is always desirable to have a power-optimized EBC concerning the false movements of the bio-robotic hand. The machine learning models can be trained to reduce the probability of misclassified movement of the EBC when at rest. This would improve significantly the classification performance of the myoelectric controller leading to a better experience for the bio-robotic hand users. In \cite{scheme2013confidence}, the authors have presented a confidence based rejection scheme for improved myoelectric classification. This method requires an additional step using the Linear discriminant analysis (LDA) log-probability outputs that results in a usable confidence score output. In this paper, we have proposed an approach to design an energy-efficient EBC. The kernelized support vector machine (SVM) has been considered as the classifier,  which is the main component of the EBC. The primary reason for choosing SVM is that they are well-suited for multiclass classification (both linear and non-linear) of various complex small-or-medium-sized dataset. We adopted the well-known $one$-$versus$-$rest$ (OvR) strategy for developing a supervised learning system to classify multiple hand gestures (a multi-class scenario) \cite{geron2019hands} based on the features, namely correlated Time-Domain Descriptors (cTDD) \cite{khushaba2016fusion} extracted from the sEMG signals as SVMs are strictly binary classifiers. However, unlike the traditional single training objective of soft margin kernelized SVM, we have formulated the training algorithm of the proposed supervised learning system as a general constrained multi-objective optimization problem (MOOP). Here, the two objectives are the minimization of the number of false negatives of a particular hand-movement class (`Rest'), and the maximization of the overall accuracy of the multiclass classifier. The presence of multiple training objectives in any learning problem will give rise to multiple optimal solutions (models), popularly known as Pareto-optimal solutions. An elitist multi-objective evolutionary algorithm (MOEA) $-$ the non-dominated sorting genetic algorithm II (NSGA-II) \cite{deb2002fast} has been used for the tuning of SVM hyperparameters (C and `gamma' ($\gamma$)  for the RBF Kernel). As we know that there are mainly two goals for multi-objective optimization, firstly, convergence to the pareto-optimal front, and secondly, diversity among the non-dominated front. In the recent past, NSGA-II has been employed to solve a wide range of complex multi-objective problems as it uses a fast non-dominated sorting procedure, an elitist-preserving approach, and a parameterless diversity-preservation mechanism.  The main reason for using evolutionary multi-criterion optimization (EMO) in our work is that it attempts to find multiple solutions with an implicit parallel search leading to faster convergence compared to any standard classical generating methods such as the normal constraint method \cite{messac2003normalized}. 

The major contributions of this manuscript are given as follows:
\begin{itemize}
\item[$\bullet$] Our work explores evolutionary computation optimization techniques to minimize the false movements of a myoelectric artificial system, eventually improving the controller's power efficiency and user experience. For this study, we have used the dataset consisting of the myoelectric signal obtained from the upper limb of different subjects using multi-electrode channels at five different limb positions, which were recorded by Khushaba $et$ $al.$ \cite{khushaba2014towards}. 
\item[$\bullet$] Moreover, Confidence-based rejection schemes are a vital point of consideration for developing sophisticated myoelectric classification systems. Existing studies \cite{scheme2013confidence} on this topic are few and usually consider a 2-step approach for addressing the issue. The proposed work integrates the optimization of the two-objective metrics, namely, the classification accuracy and false negative during the classifier's training in a single step that uses an elitist multi-objective evolutionary algorithm, the non-dominated sorting genetic algorithm II (NSGA-II) to tune the hyperparameters of the SVM classifier model. 
\item[$\bullet$] Furthermore, evaluation of the algorithms should be done carefully for a reliable generalization capability. To address this issue, we have analyzed the result on two test data sets. This provides much more flexibility to the designer in selecting the parameters of the classifier to optimize the energy efficiency of the EMG-based controller. Moreover, to the best of author's knowledge, this is the first study on sEMG signals where the multi-objective training criterion is explored to enhance the energy efficiency of the EMG-based controller.
\end{itemize}

In the remainder of the paper, we introduced various aspects of upper arm-movements classification using sEMG signal, which includes the description of a dataset (which has been used for this study) of sEMG signals, which were recorded by Khushaba $et$ $al.$ \cite{khushaba2014towards}, and details of different hand movements (classes), baseline supervised learning system based on the SVM model, the primary motivation of considering multiple training objectives to design a power-efficient EBC, and the existing related works in the literature in Section II.
Next,  we describe the proposed evolutionary multi-objective optimization approach to SVM design in detail in Section III. Section IV presents the experimental setup and the experimental results, evaluated on different subjects. Section V presents our discussion. Finally, we outline the conclusions of this paper in Section VI.
\section{Problem definition and Background}
This study uses a supervised learning framework for the multi-class classification of hand movements using sEMG signals. We utilized the dataset which were developed by Khushaba $et$ $al.$. This dataset consists of the myoelectric signals recorded in a non-invasive way from the surface of the upper limb of eleven subjects via seven electrode channels at five different limb positions \cite{khushaba2014towards} \footnote{https://www.rami-khushaba.com/electromyogram-emg-repository.html}. The following eight classes of hand movements are considered: wrist flexion ($C_1$), wrist extension ($C_2$), wrist pronation ($C_3$), wrist supination ($C_4$), power grip ($C_5$), pinch grip ($C_6$), open hand ($C_7$), and rest ($C_8$). The rationale for choosing this dataset is that it could be used to design an EMG-based control scheme for both the prosthetic hand controllers for trans-radial amputees as well as various EMG-based smart control applications such as muscle-computer interfaces for gaming. It is to be noted that the difference between developing an EMG-based smart controller for a healthy patient and an amputee will be revolve around the scope of positioning the recording electrodes on the effective muscular region. The rest of the workflow is similar irrespective of the type of subject due to the subject-specific nature of the problem.
\subsection{Supervised Learning Framework based on SVM}
In supervised learning settings, we deal with a given training dataset: a collection of labeled examples $\set{(x_i,y_i)}_{i=1}^{N}$. The composite representation of feature space can be given by a design matrix $X \in {\Bbb R}^{N \times D}$ where $D$ denotes the dimension of the feature vector. In a multiclass scenario, the label $y_i$ can be belonging to a finite set of classes $\set{1,2,...,C}$ where $C$ denotes the number of classes in the given dataset. The goal of a supervised learning algorithm is to use the design matrix and corresponding labels to learn a hypothesis or function $h: X \Rightarrow Y$, so that $h(x)$ is a good predictor for the corresponding value of $y$. In this work, we have used kernelized SVM, which is a powerful and versatile machine learning model when considering training on a non-linear dataset. The main goal of SVM is to have the widest possible margin between the decision boundary that separates the two classes and the training instances. When performing the soft margin classification, the SVM tries to find a good balance between keeping the width of the margin as large as possible and limiting the margin violations \cite{bishop2006pattern}. The soft margin classifications are fundamentally convex quadratic optimization problems with linear constraints. The optimization goal of an SVM classifier with an RBF kernel can be expressed as the following dual form after transforming the constrained optimization objective into an unconstrained one \cite{boyd2004convex}: 
\begin{equation}
\begin{aligned}
\min_{\alpha} \quad & \frac{1}{2}\sum_{i=1}^{m}\sum_{j=1}^{m}{\alpha^{(i)}\alpha^{(j)}t^{(i)}\exp (-\gamma ||\mathbf{x}^{(i)}-\mathbf{x}^{(j)}||)^2}-\sum_{i=1}^{m}\alpha^{(i)}\\
\textrm{s.t.} \quad & \alpha^{(i)}\geq0 \hspace{1.5mm} for \hspace{1mm} i=1,2,...,m    \\
\end{aligned}
\end{equation}
where $\mathbf{x}$ denotes the feature vector of a training instance. The variables $\alpha^{(i)}$ are called $Karush$-$Kuhn$-$Tucker$ (KKT) multipliers of the generalized $Lagrangian$. $\xi^{(i)}$ are known as $slack$ variable which is introduced to take into account the soft margin objective. $m$ denotes the number of training instances and the variables $t^{(i)}$ are used to define inequality constraints. Using a standard Quadratic Programming (QP) solvers, we can estimate $\hat{\alpha}^{(i)}$ which minimizes Eq.(1), and subsequently, the feature weight vector $\mathbf{w}$ and the bias term can be computed as follows \cite{geron2019hands}:
\begin{equation}
\begin{aligned} 
\hat{\mathbf{w}} &=\sum_{i=1}^{m}{\alpha^{(i)}t^{(i)}x^{(i)}} \\
\hat{b} &=\dfrac{1}{n_s}\sum_{\substack{i=1,\\ \hat{\alpha}^{(i)}\geq0}}^{m}{(t^{(i)}-\hat{\mathbf{w}}^{T}\mathbf{x^{(i)})}}
\end{aligned}
\end{equation}       
where $n_s$ denotes the number of support vectors. During the inference, an SVM binary classifier can output the distance between the test instance and the decision boundary, which can be treated as the confidence score (CS) of a given prediction. Moreover, to deal with the multi-class scenario of our application, one-vs.-rest (also known as one-vs-all)  strategy has been adopted. In this method, a binary model is learned for each class that tries to distinguish that class from all other classes, resulting in as many binary classifier models as there are classes. Hence, for the $N$-class instances dataset, we have to generate $N$-binary classifier models \cite{geron2019hands} as shown in Fig. 2. To make a prediction, all binary SVM classifier models are run on a test point. The classifier that has the highest confidence score decides what would be the predicted class label. 
%------------------------------------------------------------
\begin{figure*}[h]
  \centering
  \includegraphics[scale = 0.4]{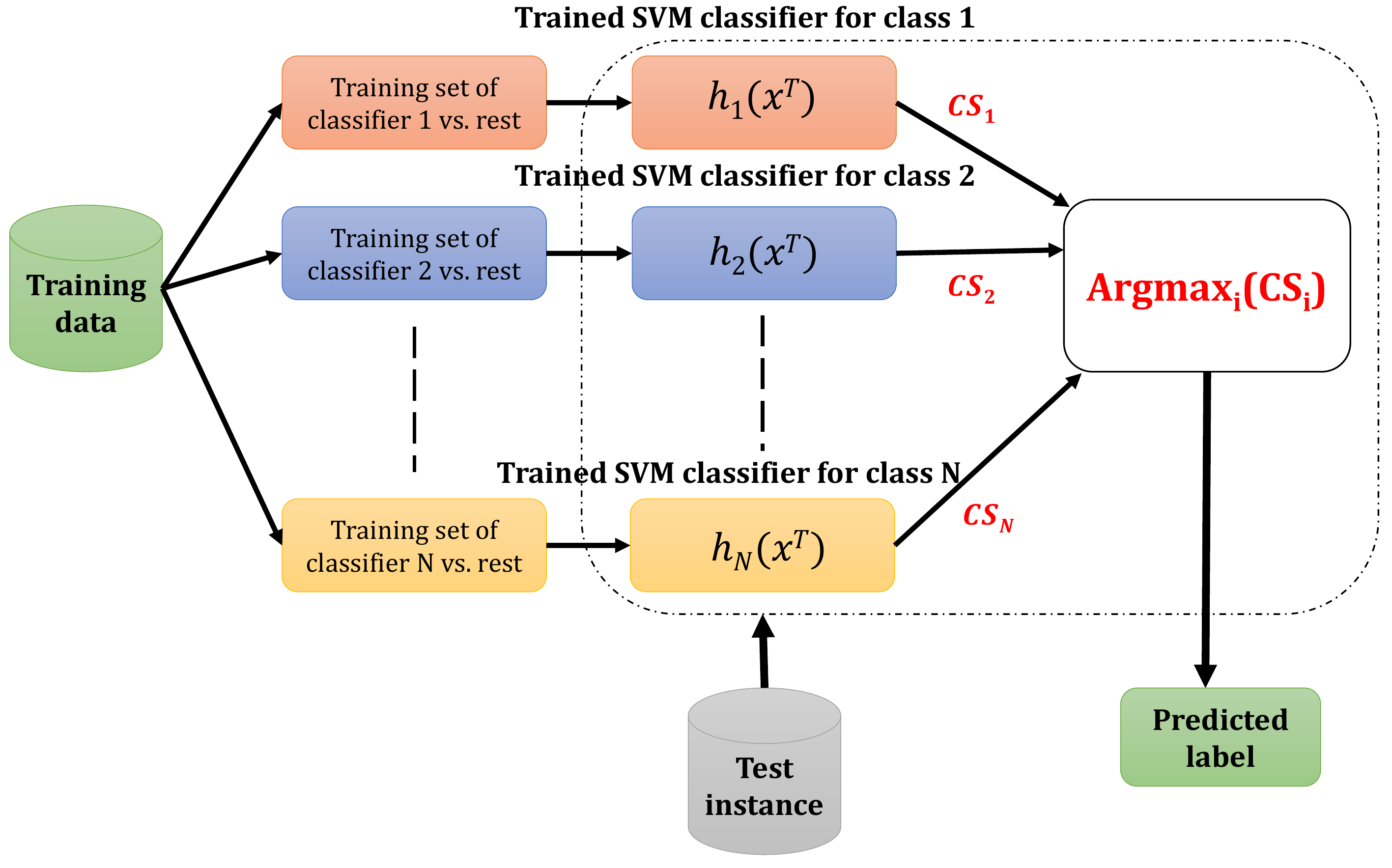}
  \caption{Multi-class Classification based on SVM: one-vs.-rest (OvR) strategy}
  \label{fig2}
\end{figure*}
%------------------------------------------------------------
\subsection{Need for multi-objective optimization: Power-efficient EBC}
The prediction result obtained from the SVM-based classifier model (in the EBC) will be utilized by the subsequent subcomponents namely, the actuator and the bio-robotic hand. To understand, the main design approach of a power-optimized controller, it would be interesting to look at the metric named $Recall$ (also called sensitivity) of $C_8$ (`Rest' position). It can be computed by analyzing the confusion matrix (as shown in Fig. 3) based on the `Rest' class ($C_8$) versus the other movement classes ($C_1$ to $C_7$); considering the rest class to be positive and the other movement classes to be negative. Here, False Negative (FN) means that the controller will perform any of the seven movements when the user wants to remain at `Rest' ($C_8$). This would lead the subsequent actuator and bio-robotic hand to take an unnecessary and undesirable (from the user perspective) action to move the prosthetic hand as the prediction result of the EBC-classifier is the main deciding factor that drives the actuator. Therefore, the model which has a low FN will be suitable for power-optimized controllers. On the other hand, FP implies that the controller will remain at rest even when the user intends to move. In this case, though the result is false the controller will be in the OFF mode. Therefore, the recall of $C_8$ should be high for avoiding unnecessary movement of the bio-robotic hand.  Hence, in this work, we set two training objectives for the multi-class SVM model: (i) minimization of the number of false negatives of $C_8$ (or equivalently, maximization of recall of $C_8$), and (ii) maximization of overall classification accuracy considering the multi-class scenario. To this end, we have employed an elitist multi-objective evolutionary algorithm to tune the hyperparameters of the SVM model.   
%------------------------------------------------------------
\begin{figure}[h]
  \centering
  \includegraphics[scale = 0.3]{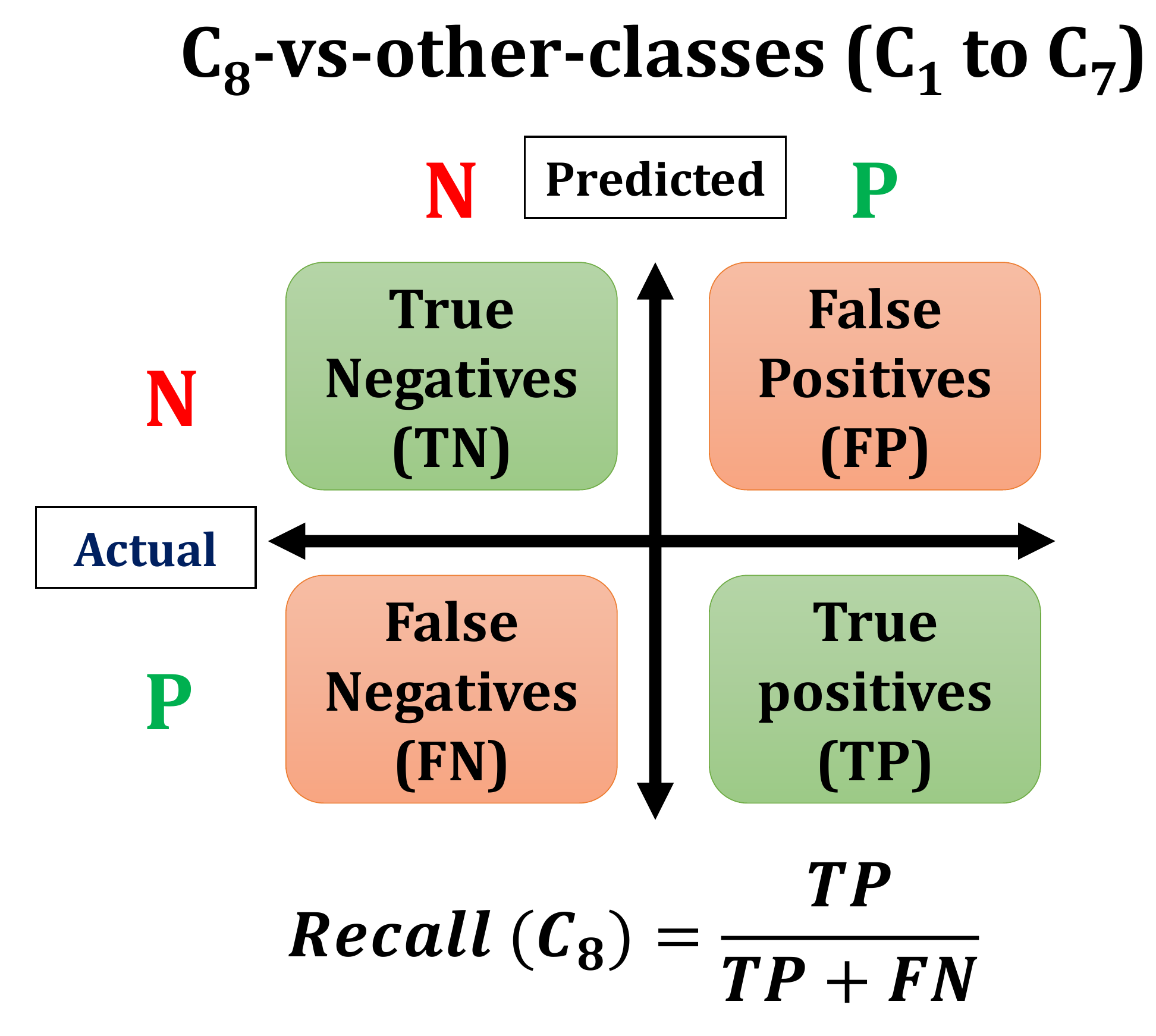}
  \caption{An illustrated confusion matrix of one of the binary classifiers}
  \label{fig3}
\end{figure}
%------------------------------------------------------------
\subsection{Related works}
Myoelectric pattern recognition for extracting the information of sEMG signals to infer the underlying muscle movements is a well-explored research area. One of the vital aspects is limiting energy consumption for power-constrained wearable devices. Researchers have tried to address this issue by reducing the computational complexity involved in the different stages of the end-to-end system. In \cite{koteshwara2018incremental}, the authors have focused on optimizing the feature bit precision for reducing the computations, applicable for designing low-power feature extraction units for seizure detection using EEG signal classification. In \cite{mukhopadhyay2018classification}, a spiking neural network based classifier model was proposed for limiting the computations involved during the inference. In \cite{kaufmanetal}, authors have studied electromyographic signal classification using evolvable hardware (EHW) classifiers which aid in self-adaptation, fast training, and compact implementation. The performance of the EHW was found to be comparable to conventional ML models such as a k-nearest neighbor, decision trees, artificial neural networks, and support vector machines. This reinforces the fact that both performance and compact implementation of classifier are important for designing wearable AI systems.

Optimizing different parameters involved in the EMG-based system is another crucial point to be considered. For instance, a genetic algorithm (GA) is utilized to determine the optimum window size and sliding interval for EMG signals in \cite{relatedworks2019}. This selection of appropriate window size is a necessary parameter that contributes to the performance of the EMG classifier. Multi-Objective Optimization Genetic Algorithm (MOOGA) is applied in \cite{relatedworks2017} to optimize the onset detection algorithms' results. A personal best-guide binary particle swarm optimization (PBPSO) for the feature selection process is proposed in \cite{too2019emg}. A feature vector of a smaller dimension reduces the complexity of the subsequent classifier model. In \cite{relatedworks2012}, the authors have used GA to optimize the number of hidden nodes and LMS algorithms to optimize the parameters, which will avoid an excessive number of hidden nodes in the NN classifier. A genetic algorithm-based SVM (GS-SVM) is employed in \cite{rong2013classification} to distinguish between the muscle states representing muscle fatigue and maximum voluntary contraction (MVC).

Various discriminative models based on linear discriminant analysis \cite{zhang2013adaptation}, SVM classifiers \cite{khushaba2014towards}\cite{mukhopadhyay2020forearm}, gradient boosting machines \cite{samui2020extreme}, and deep learning architectures \cite{mukhopadhyay2020experimental}\cite{jabbari2020emg}\cite{simao2019emg}\cite{atzori2016deep} have been successfully used for different applications in the recent past. 
%-----------------------------------------------------------------------------------
\begin{algorithm*}[htp!]
\caption{ECMO algorithm based on NSGA-II}
\scriptsize
\renewcommand{\arraystretch}{1.5}
\KwIn{Solution representation of real-coded genetic algorithm; \\ decision variable space and two objective functions: FN of class 8 and Accuracy;\\ variable bounds or constraints; \\ maximum allowed generations: $N_{max}$}
\KwOut{Pareto-front of two objective functions}
\textbf{Initialization:} Initialize the generation counter $n$:= 0; \\ create the first parent population $P(n)$ for NSGA-II using hand-picked hyperparameter values (eg: C and kernel order in SVM)\\  
\textbf{Evaluate $P(n)$:} Assign rank using dominance depth method and crowding distance operator to $P(n)$\\ 
\While{n $\leq$ $N_{max}$} {
Train SVM model for each member (set of hyperparameters) of the population
and, calculate and store two performance metrics (Accuracy and  False Negatives) \\
Using the above performance metrics sort the population into fronts using the Non-dominated Sorting Approach. The Crowding Distance is calculated for each front. \\
$M(n)$ := Crowded-tournament-selection-operator ($P(n)$); \hspace{2 cm} $\%$ Survival of the fittest   \\    
$C(n)$ := SBX-Crossover-operator ($M(n)$;\hspace{3.7 cm} $\%$ Crossover\\
$Q(n)$ := Polynomial-Mutation-operator ($C(n)$);\hspace{3.1 cm} $\%$ Mutation\\
$\hat{P}(n)$:= $P(n)\cup Q(n)$ ; \hspace{5.4 cm} $\%$ Generate merged population \\
$P(n+1)$ := Survivor-operator ($\hat{P}(n)$);\hspace{4 cm} $\%$ Using concept of Non-dominated sorting and Crowding distance metric\\
$n$ := $n+1$; } 
  	
\Return The final parent population including the pareto-optimal solutions;
\end{algorithm*}
%-----------------------------------------------------------------------------------
\section{Evolutionary Multi-objective optimization approach to SVM design}
As discussed in the preceding sections, the current work deals with a bi-objective optimization problem to select a suitable SVM model to classify multiple classes of hand movements. The performance of the SVM for any kind of pattern recognition task relies on the optimal selection of hyperparameters. Hence, the hyperparameters such as $C$ and $\gamma$  are the decision variables of the defined optimization problem. Our task would be to search for an optimal set of hyperparameters (decision variables) that would minimize or maximize the aforementioned objective functions subjected to satisfy certain constraints. In other words, the defined MOOP can be considered an SVM model selection problem in which the objective functions are implicitly defined black-box simulation models. By observing the efficiency of evolutionary computation (EC) techniques in various complex MOOPs, we have employed a population-based meta-heuristic approach that can generally find out multiple Pareto-optimal solutions in one run. Unlike classical methods such as $\epsilon$-Constraint \cite{ngatchou2005pareto}, Benson's method \cite{benson1998hybrid}, and Tchebycheff's method \cite{miettinen2012nonlinear}, EC-based techniques solve any MOOP by assigning the fitness to each solution \cite{deb2014multi}. Moreover, these techniques have a better ability to generate all the Pareto-optimal solutions, even in the case of a non-convex objective space. Fundamentally, the EC-based multi-objective optimization approach finds out the solutions in two steps as follows:
\begin{itemize}
\item Step A: Generate multiple optimal solutions. The primary goals of this step are to find solutions that should converge to the true (global) Pareto-optimal front and preserve diversity in the estimated Pareto-optimal front.
\item Step B: Select one solution using higher-level information (preference). It implies that it requires a trade-off between the objectives among the solutions.        
\end{itemize}
In this work, we have used an algorithm called EC-based multi-objective optimization (ECMO) for SVM model selection. As discussed in Section I, the primary goals in finding solutions of this  multi-objective optimization problems are mainly two: (i) the solutions should converge to the pareto-optimal front, and (ii) they should be as diverse as possible. The achievement of one goal would not necessarily help to achieve the other goal. Our proposed algorithm is based on NSGA II, one of the benchmark multi-objective EC techniques that employ a fast non-dominated sorting approach to compute the dominance-based ranking for better convergence and uses crowding distance measures to preserve the diversity. The following few definitions pertaining to the dominance and Pareto-optimality would help to describe the proposed algorithm.
\begin{itemize}
\item \textbf{Concept of dominance:} Let $\set{f_1,f_2,f_3,...,f_j,...,f_M}$ be a set of objective functions, i.e., in our case, it would be all the performance metrics for the models. A solution $s_p$ dominates solution $s_q$, if
(i) $s_p$ is no worse than $s_q$ in all objectives ($f_j(s_p) \ntriangleright f_j(s_q) \hspace{2 mm} \forall \hspace{2mm} j=1,2,3,....,M$), and, (ii) $s_p$ is strictly better than $s_q$ in at least one objective ($f_j(s_p) \triangleright f_j(s_q)$  for at least one $j \in \set{1,2,3,....,M}$).  
\item \textbf{Non-dominated set:} Among a set of solutions $P$, the non-dominated set of solutions $P_{nd}$ are those that are not dominated by any member of the set $P$.  
\item \textbf{Global Pareto-optimal set:} The non-dominated set of the entire feasible search space is defined as the global Pareto-optimal set.  
\end{itemize} 
The proposed ECMO algorithm operates based on the following steps:\\
\textbf{Step 1:} It creates the first parent population for NSGA-II using hand-picked hyperparameter
values (C and $\gamma$ for the RBF Kernel). \\
\textbf{Step 2:} Next, it uses the parent population to generate a child (offspring) population equal in size to the parent population. The concepts of crossover and mutation will be used here.
\textbf{Step 3:} Set the value of generations as required and set generation-count parameter to 0. Performance
metrics can also be used as a termination condition. \\
\textbf{Step 4:} This step consists of the following sub-steps and they are executed until the termination condition is not satisfied.
\begin{itemize}
\item Train your model for each member (set of hyperparameters) of the population and, calculate and store the performance metrics such as accuracy and false negatives.
\item Using the above performance metrics sort the population into fronts using the Non-dominated Sorting Approach. The fronts contain indices of members within the population, this helps in locating the member in the population or its performance metrics.
\item The Crowding Distance is calculated for each front.
\item The new parent population is now chosen from the fronts (starting from the
first front) using the crowded binary tournament selection method until it is the same size as the previous parent population.
\item The new child population is created using crossover and mutation.
\item The parent population is merged with the offspring population and sorted into different fronts according to the non-dominated sorting approach. The Crowding Distance is calculated for each front.
\item A population size of $N$ is selected based on their rank and crowding distance. If an entire front cannot be taken, members with a greater crowding distance are chosen first for the parent population.
\item Next, the next generation will be started with the newly selected population from the previous steps. 
\end{itemize}
\textbf{Step 5:} The final parent population is returned. \\
The pseudo-code of the composite algorithm is presented in Algorithm 1. 
The different NSGA-II operators of the proposed algorithm are described briefly in the following:
\begin{itemize}
\item Non-dominated Sorting:
It is used to place solutions on their respective fronts based on their performance. We can assign a rank to each solution using the dominance-depth method. This method uses the concept of dominance. It iteratively compares each solution in the population and all the non-dominated solutions are sorted on various fronts like front 1, front 2, front 3, and so on. However, the worst-case complexity of this dominance depth method is $O(MN^3)$. In order to avoid this expensive sorting approach, we have used a fast non-dominated sorting approach according to the NSGA-II which will require $O(MN^2)$ computations, where $M$ and $N$ denote the number of objectives and the number of members in the population, respectively. This forms the crux of NSGA-II. The storage requirement of the sorting approach is $O(N^2)$. For getting more insight into the computational steps of non-dominated sorting, please refer to \cite{deb2002fast}\cite{deb2014multi}.
\item Crowding Distance: It is calculated for each front of the population.
For every performance metric $f$:\\
a) The front is sorted using $f$ as the key.\\
b) For every member at index $i$ in the sorted front, the crowding distance (CD) is calculated as follows:
\begin{equation}
CD(i)=\dfrac{f(i+1)-f(i-1)}{max(f)-min(f)}
\end{equation}
where $max(f)$ and $min(f)$ are the maximum and minimum value of performance metric $f$ for the entire population.\\
c) The members at the first and last position of the sorted front are assigned a
distance of infinity $(\infty)$ or a very large number. 
\item Crowded binary tournament selection: This operator is used to choose good (above-average) solutions. This tournament selection is accomplished on a pool of $N$ solutions whose ranks and crowding distances are already computed. It operates as per the following two steps:\\
(a) At first, two solutions are randomly chosen from the population. \\
(b) Then, out of these two solutions, a particular solution is selected based on the following criterion:\\
case-I: If the rank of the two solutions are different, then the solution with the better rank is selected.\\
case-II: If the rank of the two solutions is the same, then the solution with a larger crowding distance will be get selected. \\
case-III: If the rank, as well as the crowding distance of the solutions, are the same, then any solution is randomly selected. \\
After each tournament  $\frac{N}{2}$ number of solutions are got selected. Hence, we perform this tournament selection two times in each generation to end up with a good quality parent population with $N$ members.    
\item Crossover: Inside the loop of each generation, this is the first variation operator after selection. Crossover is responsible for creating new solutions. These new solutions explore the search space. The desired properties of a standard crossover operator are as follows: (a) crossover operator should not steer the search in any particular direction as it does not utilize any fitness values. To this end, it should generate an offspring population such that the mean of the offspring population should be the same as that of the parent population. However, the variance may change during this process. (b) Secondly, the crossover operator should increase the population diversity by increasing the variance of the population. Otherwise, it may lead to a premature convergence. The proposed ECMO has used a simulated binary crossover operator (SBX) to generate the offspring population. The crossover is performed with probability ($p_c$). The SBX is based on a non-linear probability distribution function \cite{deb1995simulated} as given below:
\begin{equation}
{p}(\beta) = \begin{cases} 0.5(\eta_c+1)\beta^{\eta_c}, & \beta \leq 1 \\ 0.5(\eta_c+1)\beta^{\frac{1}{\eta_c+2}}, & otherwise  \end{cases}
\end{equation}
where $\eta_c$ is an empirical parameter which is set during simulation.  
\item Mutation: It is the second variation operator which operates on the offspring population generated from the crossover stage. Its main purpose is to exploit the search space by perturbing the solutions. Our proposed method used a polynomial mutation operator \cite{deb1995simulated} for perturbing any solution. The time and space complexities of various operators are shown in Table I. 
\end{itemize}   
\begin{table}[]
\centering
\scriptsize
\caption{Time and Space complexity}
\renewcommand{\arraystretch}{1.5}
\begin{tabular}{||c|c|c||}
\hline \hline
\textbf{NSGA component} & \textbf{Time complexity} & \textbf{Space complexity} \\ \hline
Non-dominated Sorting   & $O(MN^2)$                   & $O(N^2)$                      \\ \hline
Crowding Distance       & $O(MNlogN)$                    & $O(N)$                      \\ \hline
Crossover               & $O(1)$                     & $O(1)$                      \\ \hline
Mutation               & $O(1)$                     & $O(1)$                      \\ \hline 
\hline
\end{tabular}
\end{table}
%-----------------------------------------------------------------------
 \begin{figure*}[htp!]

 \centering

  \begin{tabular}{cc}

    \includegraphics[scale =0.65]{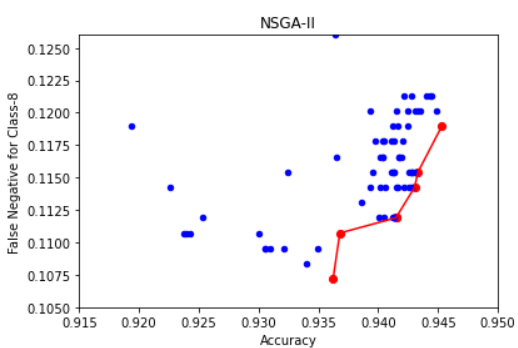} &
    
    \includegraphics[scale =0.65]{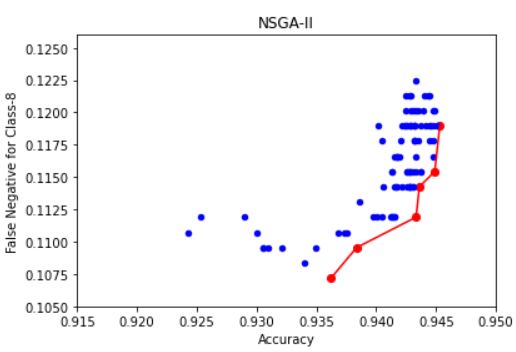} \\ 
    
    \includegraphics[scale =0.65]{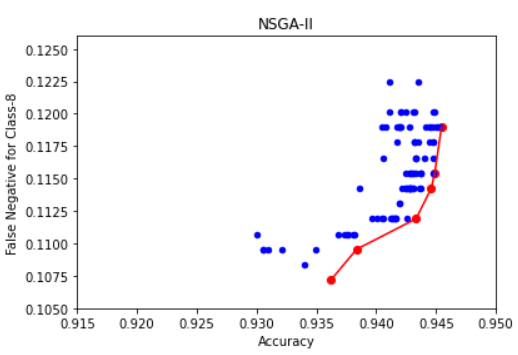} &
    
    \includegraphics[scale =0.65]{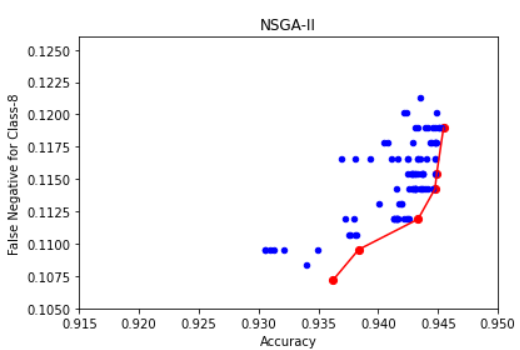} \\ 
    
    \includegraphics[scale =0.65]{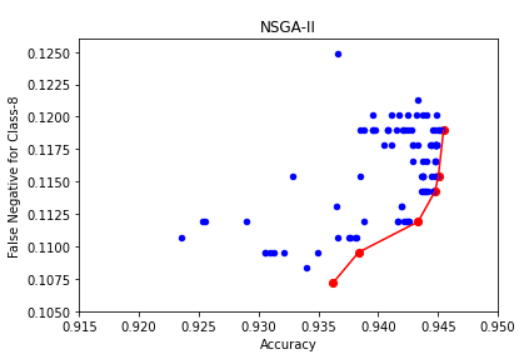} &
    
    \includegraphics[scale =0.65]{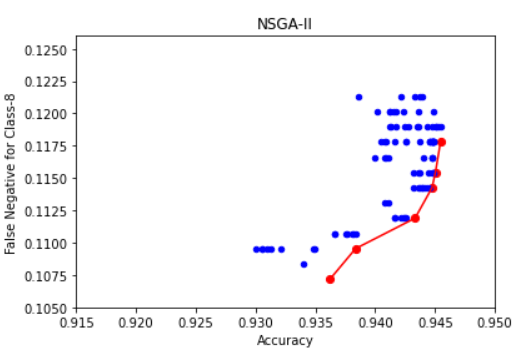} \\
    
    \includegraphics[scale =0.65]{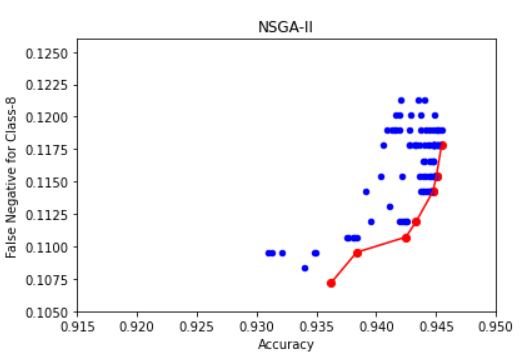} &
    
    \includegraphics[scale =0.65]{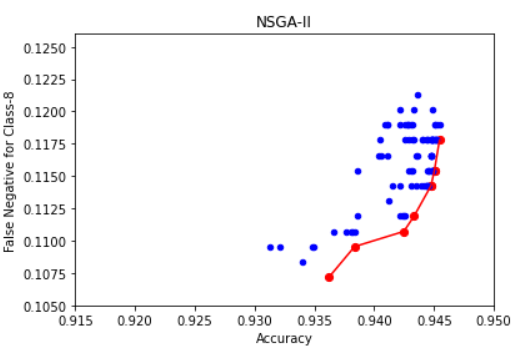} \\
    
    \includegraphics[scale =0.65]{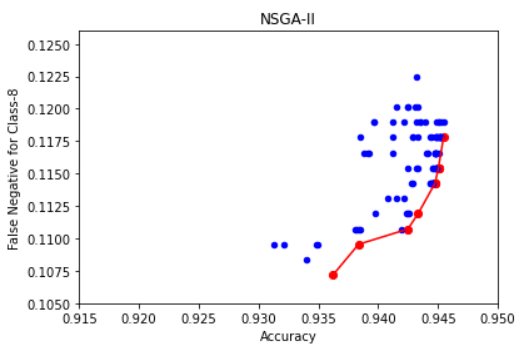} &
    
    \includegraphics[scale =0.65]{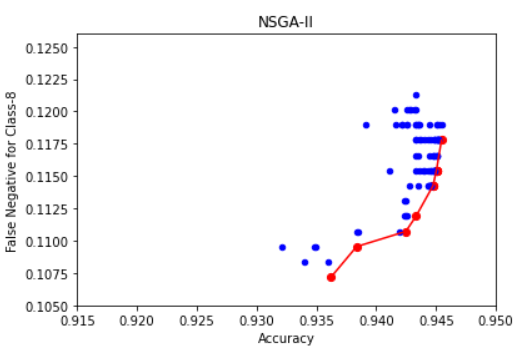} \\   

  \end{tabular}
  \label{fig4}\caption{NSGA-II Plots (Accuracy vs. FN for class 8) for Subject 10}
\end{figure*}
%--------------------------------------------------------------  
\section{Experimental results}
As discussed in Section II, our experimental setup was configured to handle the multi-class (number of classes = 8)  classification problem. Furthermore, in our setup, we had to take care of two more aspects in general:
\begin{itemize}
\item Firstly, the EMG data is absolutely subject-specific. It implies that we had to deal separately with each subject taken from the dataset, and subsequently, different dedicated models have been trained and optimized for different subjects. In fact, all the performance measures mentioned in our analysis will be the average score considering the eleven subjects (nine male and two females). 
\item Secondly, the limb position invariance is a vital aspect of the entire design process of this type of EMG based controller. It is always a convenient aspect from the user's perspective that if we can train a model using the features extracted from the EMG data collected from a lesser number of limb positions while anticipating that the trained model will be able to show significantly good accuracy even if the test data comes from a different position (unseen during training). Based on the analysis shown in \cite{khushaba2014towards} as well as in our previous work \cite{mukhopadhyay2020experimental}, we have trained the classifiers using the sEMG data taken from positions 1(P1), 3(P3), and 5(P5). To evaluate our system, two test sets: TS1 and TS2 have been created. TS1 is completely disjoint from the train set as it contains the data only from completely unseen positions (P2 and P4). It would help to validate the generalization capability of the proposed system while implementing the same movements at a different limb position. On the other hand, to capture a real-time testing session, TS1 includes data from all the limb positions (P1 to P5), including those completely unseen during training\footnote{Our implementation of these experiments will be openly available online after acceptance of our paper.}. 
\end{itemize}
The data distribution for each subject is shown in Table II. The data samples are randomly chosen from the 5 seconds time period and assigned to the train/validation/test set so that the eight possible classes are distributed in a balanced manner.  Furthermore, for 8 different type of movement classes and 6 trials for a particular limb position, the number of datasamples is 9456 (8 classes $\times$ 6 trials $\times$ 1 limb position $\times$ 197 datasamples). Hence, our train set consists of data-points belonging to three positions i.e. total 28368 (=9456 $\times$ 3). The TS1 consists of 12168 data-points (1521 data-points per class) belonging to two limb positions P2 and P4. The TS2 consists of 7040 data-points (880 data-points per class) belonging to all the possible five limb positions (P1 to P5).

The cTTD feature descriptors were computed on the data samples. cTTD is a set of six feature coefficients ($f_1$-$f_6$) namely root squared zero, second, and fourth order moment($f_1$-$f_3$), sparseness ($f_4$), irregularity factor ($f_5$), waveform length ($f_6$) \cite{khushaba2016fusion}. This set of feature coefficients contain a correlation of time domain descriptors with a non-linear version of time domain descriptors capturing the frequency domain information, making it both robust and computationally cost effective as discussed previously by Khushaba \cite{al2016improving}.  
\begin{table}[]
\centering
\scriptsize
\renewcommand{\arraystretch}{1.5}
\caption{Data distribution for each subject}
\begin{tabular}{||c|c|c||}
\hline \hline
\textbf{Subset}     & \textbf{Number of data-points} & \textbf{Positions}        \\ \hline
\textbf{Train}      & 28368                      & P1, P3 and P5             \\ \hline
\textbf{Test set 1} & 12168                      & P2 and P4                 \\ \hline
\textbf{Test set 2} & 7040                       & All the positions (P1-P5) \\ \hline \hline
\end{tabular}
\end{table}

\begin{table}[]
\centering
\caption{Various empirical parameters}
\scriptsize
\renewcommand{\arraystretch}{1.5}
\begin{tabular}{||c|c||}
\hline
\hline
\textbf{Parameters}      & \textbf{Value} \\ \hline
Population               & 98             \\ \hline
Total generation count   & 10             \\ \hline
Probability of crossover & 0.6            \\ \hline
$\eta_c$ for SBX  		 & 15            \\ \hline
Probability of mutation  & 0.4            \\ \hline
$\eta_m$ for polynomial mutation  & 20            \\ \hline \hline
\end{tabular}
\end{table}

%-----------------------------------------------------------------------
 \begin{figure*}[htp!]

 \centering

  \begin{tabular}{cc}

    \includegraphics[scale =0.5]{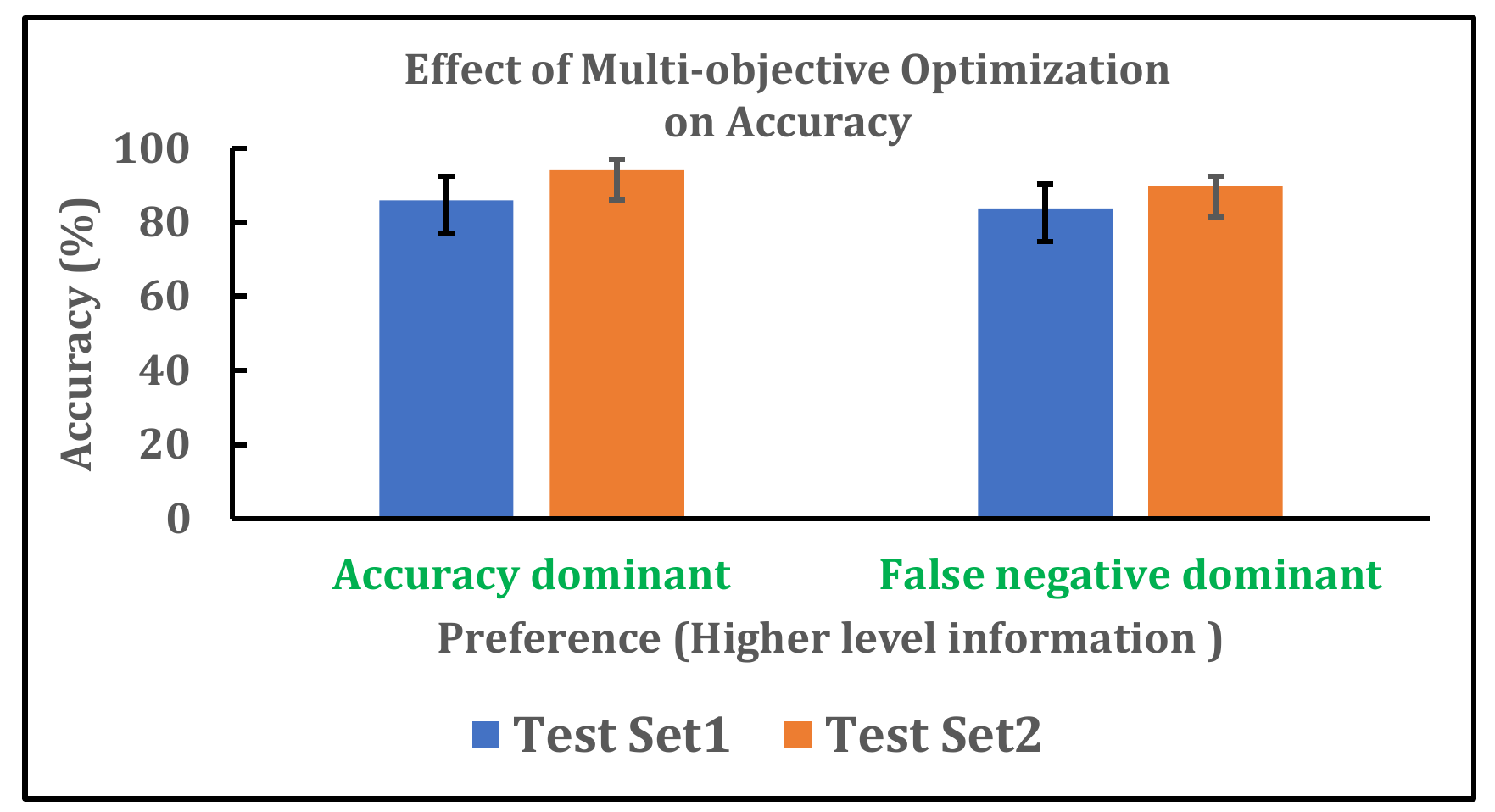} &
    
    \includegraphics[scale =0.5]{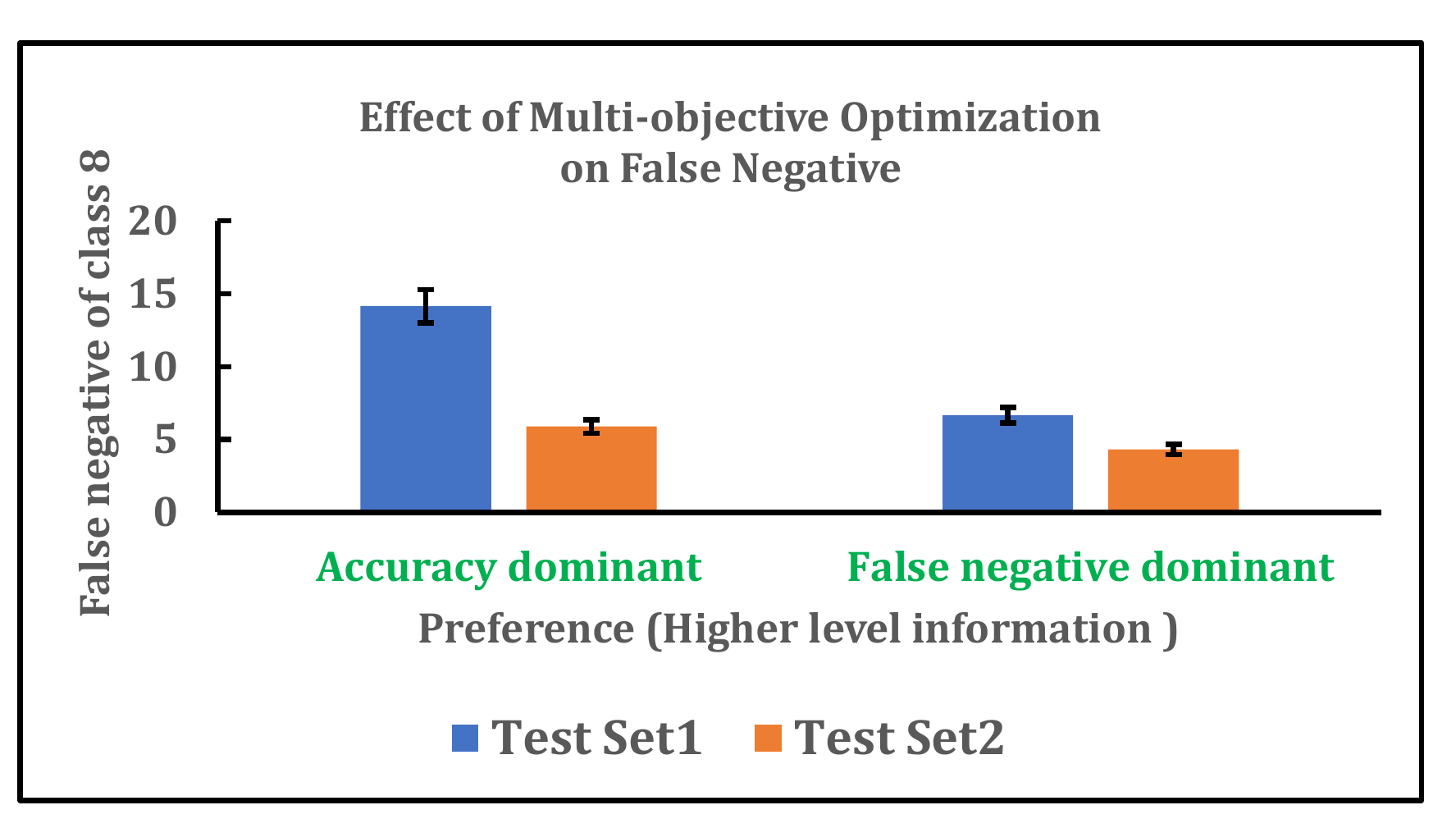} 

  \end{tabular}
  \label{fig10}\caption{Effect of Multi-objective Optimization
on Accuracy and FN of class 8}
\end{figure*}
%---------------------------------------------------------------------------
The NSGA-II based ECMO algorithm has been used for the tuning of SVM hyperparameters (C and
$\gamma$ for the RBF Kernel) for each subject. The training set and a hold-out validation set (20\% of training data) are used in the first stage of NSGA-II. An OvR SVM model is trained using the train Set for every member of the population (set of hyperparameters). We calculate Accuracy and False Negative for Class 8 on TS2 for each SVM model. These serve as the performance metrics for NSGA-II. After NSGA-II has been completed, we select the solution with the highest Accuracy and the solution with the lowest FN for Class 8 from the non-dominated front of the last generation of NSGA-II. These solutions are simply sets of hyperparameters used for SVM. These selected solutions are once again used to train OVR SVM models, these models are validated using TS1. The objectives used for optimization are Accuracy and False Negative for the rest class (Class 8). The initial population for NSGA-II consists of some handpicked sets of hyperparameter values spread out over the search space. An SVM model is trained for each set of hyperparameters. The Accuracy and False Negative scores of each model on the validation set are used for non-dominated sorting. The algorithm is run for 10 generations for each subject, and the best performing solutions based on non-dominated sorting and crowding distance are taken as the parent population for the next generation. Each member of the population is a set of 2 hyperparameters needed for SVM. Two members of the parent population are chosen at random for creating a member of the child population. The SBX and polynomial-time mutation operator are used in the variation process as discussed in Section III. The parameter $\beta$ is calculated by equating the probability curve (dictated by Eq. (4)) to a randomly generated number $u$ from a uniform probability distribution, i.e., $u \in [0,1]$. All the essential parameters of the implemented ECMO algorithm are shown in Table III. Through random mutation and crossover, the algorithm reaches the optimal solutions. The solution with the highest Accuracy Score and the solution with the best False Negative Score are tested on a disjoint test set. An instance of NSGA-II plots on the objective space (accuracy vs. false negative of class 8) for Subject 10 is shown in Fig. 4. It is worth mentioning that, a non-dominated front is created in each generation. At the end of the last generation, it can be observed that the Pareto-optimal front has been estimated. Hence, the subsequent process is to impose a preference on the objective space to select a particular solution from the estimated Pareto-optimal set. Based on the set preference, we can categorize the solutions into two groups: (i) Accuracy dominant solutions, and (ii) false-negative dominated solutions. The performance of the models obtained from the estimated Pareto-optimal set is evaluated for each of the individual subjects. The mean value of the metrics averaged over the eleven subjects has been depicted in Fig. 5. Both accuracy dominant and FN dominant categories have shown superior performance on Test Set 2 as this set contains the signal instances from all the possible positions. Furthermore, we can notice that the accuracy levels are similar for both accuracy dominant solutions and false-negative dominated solutions as seen in Fig. 5(a) when tested on both the datasets. However, there is a considerable (almost 50\%) reduction in misclassification error due to false negative for the FN dominant case (please refer to Fig. 5(b)), validating our objective towards building a robust and user-convenient prosthesis system.
\section{Discussion}
The main purpose of this study is to bring in the multi-objective training criterion to reduce the false movements at a particular upper-limb position towards enhancing the energy efficiency of the EMG controller. However, our implementation of the algorithm and evaluation method has not neglected the upper-limb position invariance training strategy which is now a well-established method \cite{khushaba2014towards}\cite{mukhopadhyay2020experimental}.  The proposed ECMO algorithm is based on NSGA-II. Hence, it is an iterative improvement algorithm that runs for 10 generations for each subject, the best performing solutions based on non-dominated sorting and crowding distance are taken as the parent population for the next generation. Through polynomial mutation and crossover, the algorithm generated the optimal solutions by estimating a Pareto-optimal front. 
From one generation to the next, the non-dominated front can change in the following scenarios:
\begin{itemize}
\item A new non-dominated solution is found with identical performance to a previous the non-dominated solution, the new solution will be a part of the non-dominated front. The shape of the front remains the same. 
\item A new non-dominated solution is found which dominates one or more members of the previous non-dominated front. The new solution is a part of the non-dominated front and the dominated solutions are not present in the new non-dominated front. This change is observed from Generation 1 to Generation 2 and from Generation 2 to Generation 3.
\item A new non-dominated solution is found that neither dominates nor is identical to any member of the previous non-dominated front. The new solution will be a part of the non-dominated front. This change is observed from Generation 6 to Generation 7.
\end{itemize}
Additionally, we see that the solutions move towards the non-dominated front with each
generation and there are fewer outliers. The NSGA-II non-dominated solution with the best Accuracy score and the solution with the lowest False Negative score is tested on the disjoint test set (TS1) for every subject. The accuracy dominant solutions have a better accuracy on the disjoint set. Similarly, the false negative dominant solutions have a better false negative score on the disjoint set. A trade-off can be made by choosing other solutions in the non-dominated front obtained from the proposed ECMO algorithm.
\section{Conclusions}
This study has applied the concept of Pareto-optimality to an SVM-based sEMG signal classification system. We have identified a specific multi-objective optimization problem in order to enhance the power efficiency of a myoelectric controller for applications pertaining to upper-limb prostheses and bio-robotic hand movement systems. Our proposed design strategy is aimed at reducing the number of false negatives of a particular hand-movement class while at the same time, it would take care of minimizing a cost function that accounts for the classification error on a given training data. From the viewpoint of multi-objective optimization, it is quite evident that there cannot be a single learning model that can satisfy different objectives at the same time. In this sense, Pareto-based multi-objective optimization is the only way to deal with the conflicting objectives in a supervised learning framework. To this end, we have proposed an algorithm based on a standard evolutionary multi-objective optimization method to solve the Pareto-optimality in model selection. The extensive evaluation on the two subsets of unseen test data has proven the generalization capability of the trained kernelized SVM models. However, this work did not take into account the trade-off between classification accuracy and model complexity in model selection. Hence, a future research direction is to formulate the SVM model selection as an evolutionary multi-objective optimization problem with two objectives: (a) minimizing the size of the training set by employing data preprocessing techniques such as instant selection \cite{garcia2016tutorial}\cite{rosales2017evolutionary} to bring down the computational complexity in the model training, and (b) maximizing the classification performance attained by the selection of the training instances. Moreover, it would be also interesting in the future to explore various TinyML approaches \cite{warden2019tinyml}\cite{tinyml} to generate a memory-efficient hand gesture recognition model which can be deployed on different extremely low-power and low-cost microcontrollers for real-time hand prosthesis control \cite{bian2021capacitive}. Furthermore, the future work will be devoted to the exploration of various other types of Pareto-based multi-objective ensemble learning \cite{chandra2006trade} as well as employing multi-objective neural network optimization \cite{jin2006multi}\cite{yen2006multi} for sEMG signal classification.
 
\bibliographystyle{IEEEtran}
\bibliography{ref_moop}

\end{document}